\documentclass[letterpaper, 10 pt, conference]{ieeeconf}  

\IEEEoverridecommandlockouts                              

\usepackage{subfiles}
\usepackage{xcolor}
\usepackage{graphicx} 
\usepackage{subcaption}
\usepackage{hyperref}
\usepackage{flushend}
\usepackage[export]{adjustbox}
\usepackage{array}
\usepackage{mathrsfs}
\usepackage[font=small]{caption}
\usepackage{amsmath}
\usepackage{amssymb}
\usepackage{multirow}
\usepackage{upgreek}
\usepackage[belowskip=-15pt,aboveskip=0pt]{caption}
\usepackage{cite}
\usepackage{tabularx}
\let\labelindent\relax
\usepackage{enumitem,kantlipsum} 
\usepackage{bm}
\usepackage{physics}
\usepackage{stackengine}

\definecolor{sher}{rgb}{0.0,0.3,0.5}
\definecolor{Alirezacolor}{rgb}{1,0.1,0.3}

\title{\LARGE \bf Pre-Surgical Planner for Robot-Assisted Vitreoretinal Surgery: Integrating Eye Posture, Robot Position and Insertion Point }

\author{Satoshi Inagaki$^{1,2}$, Alireza Alikhani$^{2}$, Nassir Navab$^{2}$, Peter C. Issa$^{3}$ and M. Ali Nasseri$^{3,4}$
\thanks{This paper is partially supported by NSK Ltd. and partially by the state of Bavaria through Bayerische Forschungsstigtung (BFS) under Grant AZ-1592-23-ForNeRo. Corresponding author: M. Ali Nasseri \tt\small (ali.nasseri@mri.tum.de)}
\thanks{$^{1}$  S.Inagaki is with NSK Ltd., Japan.
}%
\thanks{$^{2}$ S.Inagaki, A. Alikhani, and N. Navab is with Department of Computer Science in Technische Universit\"{a}t M\"{u}nchen, M\"{u}nchen 85748 Germany. 
}
\thanks{$^{3}$ Peter C. Issa and M. Ali Nasseri are with Augenklinik und Poliklinik, Klinikum rechts der Isar der Technische Universit\"{a}t M\"{u}nchen, M\"{u}nchen 81675 Germany.
}%
\thanks{$^{4}$ M. Ali Nasseri is with the Department of Biomedical Engineering, University of Alberta, Canada.}
}%

\begin{document}
\maketitle
\thispagestyle{empty}
\pagestyle{empty}
\begin{abstract}

Several robotic frameworks have been recently developed to assist ophthalmic surgeons in performing complex vitreoretinal procedures such as subretinal injection of advanced therapeutics.
These surgical robots show promising capabilities; however, most of them have to limit their working volume to achieve maximum accuracy. Moreover, the visible area seen through the surgical microscope is limited and solely depends on the eye posture. If the eye posture, trocar position, and robot configuration are not correctly arranged, the instrument may not reach the target position, and the preparation will have to be redone.
Therefore, this paper proposes the optimization framework of the eye tilting and the robot positioning to reach various target areas for different patients.
Our method was validated with an adjustable phantom eye model, and the error of this workflow was 0.13 $\pm$ 1.65 deg (rotational joint around Y axis), -1.40 $\pm$ 1.13 deg (around X axis), and 1.80 $\pm$ 1.51 mm (depth, Z).
The potential error sources are also analyzed in the discussion section.
\bstctlcite{IEEEexample:BSTcontrol}


\end{abstract}

\section{Introduction}
Retinal surgery is a highly delicate medical procedure that requires exceptional precision and fine motor control from ophthalmic surgeons~\cite{10611207}. Procedures including subretinal injection, retinal vein cannulation, and retinal peeling are a few examples. In retinal vein cannulation, the surgeon must carefully insert an instrument into a vessel with a diameter of $100-400$ µm and maintain its position for several minutes~\cite{KUleventowardsaclinically}. However, natural physiological tremors in humans make achieving high precision challenging. Hence, the ophthalmic surgeries field has adopted robotic platforms to assist and enhance surgeon's capabilities during retinal surgeries~\cite{ sixdegree1993,gerber2020advanced,rahimy2013robot, nasseri2013introduction, 10340955, he2012toward}.
In ophthalmic surgery, the fundus area is observed through the cornea and the visible area is limited, therefore, the position of the visible area is adjusted by tilting the eye~\cite{KUlevensetupandmethod,Koyama,weidesign,weiperformance,haorandesign}. 
Moreover, a trocar is planted as an entry port to minimize the scleral stress on the eyeball. The surgical robot has to control the instrument around this, which is called remote center of motion (RCM) control. 
Therefore, the trocar position limits the surgical robot's working area, and the target areas vary depending on the type of operation and the surgical site shift from one surgery to another~\cite{faridpooya2022randomised}.

\begin{figure}[t]
    \centering
\includegraphics[width=1.0\linewidth]{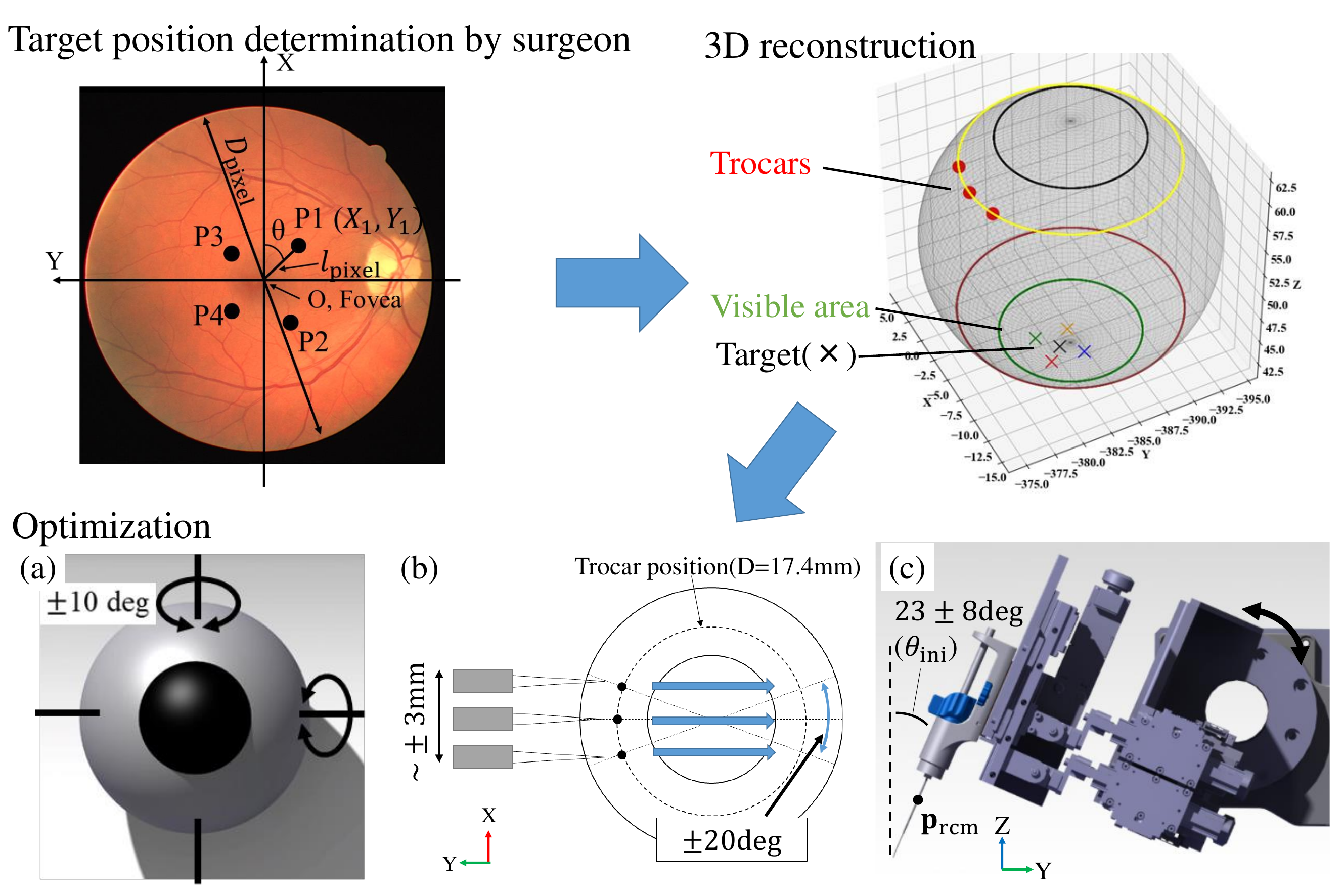}
    \caption{Overview of the surgical planner. (a)-(c) in optimization shows the strategy to expand the accessible area.} 
    \label{fig:strategy}
    \vspace{-3pt}
\end{figure}

To facilitate safe operation at the target area, this area has to be covered both by the working area of the robot and the visible area. Therefore, we conducted an accessibility analysis in our previous work~\cite{accessibilityanalysis}. The working area is shifted by the eyeball tilting, the RCM point, and the robot tilting. The eye posture shifts the visible area. Through this accessibility analysis, we defined the strategy to expand the visible-accessible area as shown in Fig.\ref{fig:strategy}(a)-(c) and confirmed it by analysis and experiment. 
Although this work proved that these strategies contribute to the expansion of the visible-accessible area, it could not step into how to optimize these factors depending on the target area. If the eye posture, trocar position, and robot configuration is not correctly arranged, the instrument may not reach the target position, and the preparation will have to be redone. In this paper, we advance this analysis and propose a preoperative planning framework that converts the target position defined by the surgeon in the 2D diagnostic image into a 3D position and optimizes the eye and robot posture and insertion position from the target position (Fig.\ref{fig:strategy}). Since this work aims to overlap the target area with the visible-accessible area, the accuracy is not so much required, assuming around 1 mm, compared to the accuracy required for subretinal injection. 

This paper's contribution is first to propose a workflow and method of surgical planning, and secondly, this workflow was experimentally validated using a realistic eye phantom. Thirdly, the multiple potential error sources are analyzed in the discussion section.

\subsection{Related Work}
There are many studies that focused on preoperative planning using the da Vinci system, assuming surgeries such as laparoscopic and cardiac surgeries\cite{davincipreoperative1,davincipreoperative2,davincipreoperative3,davincipreoperative4,davincipreoperative5,davincipreoperative6,davincipreoperative7}. These studies have proposed methods for optimizing trocar placement and the initial posture of the robot. Based on the information from the CT scan, these optimizations are proposed by considering the accessibility of the surgical robot, the endoscopic field of view, arm dexterity, arm-to-arm interference, and even obstacles such as internal organs. Other studies have used general-purpose robotic arms to optimize trocar positioning and surgical instrument mounting\cite{trocaroptimization1,trocaroptimization2,trocaroptimization3}.
However, to the best of our knowledge, there are no studies that focus on the subject of preoperative planning in the field of vitreoretinal surgery. The major difference between vitreoretinal surgery and the mentioned surgeries is that the posture of the eye can be changed, which changes the field of view and accessibility. 

As for the eye surgery, there are several studies on eye rotation.
Smits et al. developed a fixture mechanism for repositioning RCM points~\cite {KUlevensetupandmethod}. Koyama et al.~\cite{Koyama} and the group of~\cite{weidesign,weiperformance,haorandesign} have developed algorithms to perform orbital manipulation within an eyeball utlizing dual-arm robot. However, their focus was more on RCM repositioning or rotating eyeballs rather than optimizing the visible-accessible area.
Therefore, this paper is the first to address presurgical planning for robot-assisted vitreoretinal surgery. 

\section{Introduction Of The Robot} \label{sec:robotmechanics}
\begin{figure}[t]
    \centering
    \vspace{3pt}
    \includegraphics[width=0.92\linewidth]{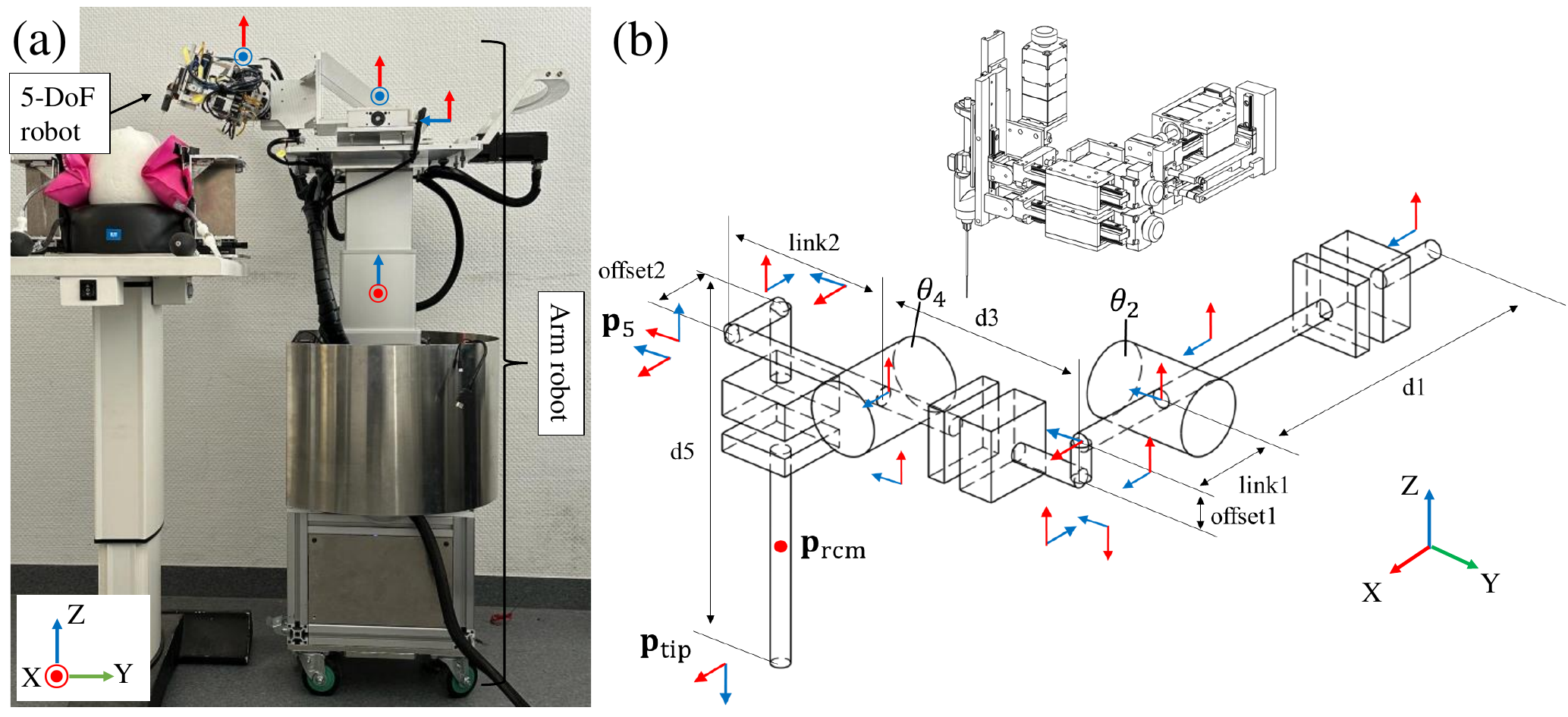}
    \caption{(a) shows the whole view of the robot. (b) shows the kinematic structure of the 5-DoF surgical robot.} 
    \label{fig:kinematics}
    \vspace{-2pt}
\end{figure}

The whole structure of our surgical robot is shown in Fig.~\ref{fig:kinematics}(a). Our robot can be divided into two central parts: the 5-DoF robot (Fig.~\ref{fig:kinematics}(b)) and the arm robot.
The 5-DoF robot is responsible for the tiny and precise movement of the surgery.
The robot consists of two parallel coupled joint mechanisms (PCJM) for translation and rotation in the X and Y axis and a decoupled prismatic joint for movement along the Z-axis. The kinematics is described in our previous work~\cite{accessibilityanalysis,PKC-RCM,nasseri2013kinematics}.
This arm robot is responsible for coarse movement.
The arm robot brings the 5-DoF robot to the RCM point and tilts it to the required angle.

\section{Method}{\label{sec:method}}




\subsection{Accessibility Calculation}\label{sec:kinematic}
The RCM point ($\vb{p}_{\rm{rcm}}$) is along the instrument and can be expressed as follows~\cite{virtualfixture}.
\begin{equation}\label{eq:RCM_control}
\begin{split}
    \vb{p}_{\rm{rcm}} = \lambda(\vb{p}_{\rm{tip}}-\vb{p}_{5})
\end{split}
\end{equation}
where $\vb{p}_{5}$ is the endpoint of the last joint of the robot and $\vb{p}_{tip}$ is the tip of the instrument as shown in Fig.~\ref{fig:kinematics}(b).

In order to calculate the accessibility, the distance between the tooltip 
 ($\vb{p}_{\rm{tip}}$) and the center of the eyeball ($\vb{p}_{\rm{center}}$) should be equal to the radius of the eyeball ($r$), (Fig.\ref{fig:eyeballtilt}).
\begin{equation}\label{eq:accessibility}
\begin{split}
    |\vb{p}_{\rm{tip}} - \vb{p}_{\rm{center}}| = r\\
\end{split}
\end{equation}

\subsection{Strategy to adjust the visible-accessible area} \label{sec:strategy}
The strategy to optimize the visible-accessible area is defined in our previous work\cite{accessibilityanalysis} as follows.
\begin{enumerate}
  \item Adjusting the eye posture within $\pm$10 deg around the X and Y axis (Fig.~\ref{fig:strategy}(a))
  \item Selecting insertion point (Fig.~\ref{fig:strategy}(b)) from three trocar. One trocar is 3 or 9 o'clock position and two trocars are planted at $\pm$20 deg position from there, which leads to about $\pm$3 mm. This helps to move the working area along the X-axis. 
  \item Adjusting the initial tilting angle of the robot within 23$\pm$8 deg (Fig.~\ref{fig:strategy}(c)). This helps to move the working area along the Y-axis.
\end{enumerate}

\subsection{Workflow} \label{sec:workflow}
The proposed workflow for surgical planning for robot-assisted vitreoretinal surgery is as follows:
\begin{description}
  \item[Diagnostic phase] Prior to starting the operation. 
  \item[STEP1] Surgeons take the fundus and OCT image and determine the target point by clicking the image.
  \item[STEP2] The system estimates the 3D position of the target after 2D-3D eyeball calculation (Function1)
  \item[Preparation phase] The preparation of the operation is started, and the robot is placed next to the patient. Three trocars for inserting the instrument are planted as defined in Sec.\ref{sec:strategy} (Fig.\ref{fig:strategy}(b)). Additionally, two trocars are planted for the light pipe and infusion.
  \item[STEP3] The system calculates the posture and the position of the eyeball and the trocars. (Function2)
  \item[STEP4] The system proposes tilting the angle of the eyeball to the surgeon by GUI. (Function3)
  \item[STEP5] The system proposes the best trocar to insert and calculate the initial tilting angle of the robot and the initial position of each axis. (Function4)
  \item[Operation phase] The instrument is inserted through the trocar and approaches the target.
  \item[STEP6] The system calculates the target angle and position to approach the target. (Function5)
\end{description}
After the workflow is completely executed, the instrument is close to the retina, and then the surgeon can adjust the instrument position using OCT. 
Each critical step within the proposed workflow is associated with a specific “function” described in detail.

\subsection{Function1. 2D to 3D reconstruction of the target position }\label{sec: targetposcal}
\begin{figure}[b]
    \centering
    \vspace{0pt}
    \includegraphics[width=1.0\linewidth]{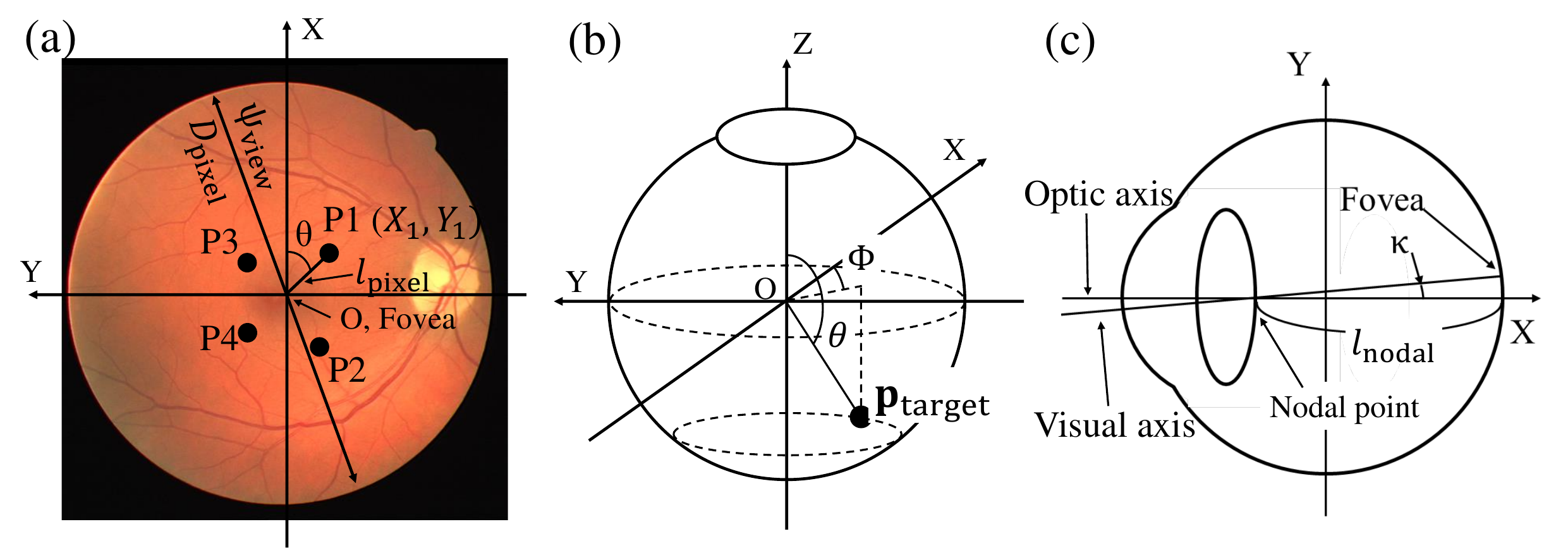}
    \caption{2D-3D conversion. (a)Diagnostic image\cite{fundusimage}. The surgeon defines the target. (b) The polar coordinate of the target is obtained from (a). (c) Schematic diagram of the optical and the visual axis. }
    \label{fig:2D3Dconvert}
    \vspace{12pt}
\end{figure}

In this section, we assume that the center of the diagnostic image is the fovea (Fig.\ref{fig:2D3Dconvert}). For this assumption, the diagnostic image has to be taken from right in front of the eye. 
A reference is required to convert the pixel to the actual length. The angle of the view of the diagnostic image ($\psi_{\rm{view}}$) is usually 45deg or 60 deg. The diameter of the image ($D_{\rm{FoV}}$) is obtained from this angle of view. Utilizing this diameter of the image, the conversion factor ($k_{\rm{pixel2mm}}$) is calculated.
\begin{equation}\label{eq:conversion}
\begin{split}
D_{\rm{FoV}} = 2r\sin(\psi_{\rm{view}}/2)    \\
k_{\rm{pixel2mm}} =  D_{\rm{FoV}}/D_{\rm{pixel}} \\
\end{split}
\end{equation}
where $r$ is the radius of the eyeball (12.1mm) and $D_{\rm{pixel}}$ is the diameter of the image in pixels. After performing several pre-processing steps, such as removing the color channel and denoising, utilizing the Ellipse Hough Transform\cite{houghcircle}, the diameter ($D_{\rm{pixel}}$) is calculated. 
The polar coordinate ($\theta$, $\phi$) of the target is obtained from the 2D position of the target ($\vb{p}_{\rm{target}} =  (x, y)$) and the  distance from the center ($l_{\rm{pixel}}$). 
\begin{equation}\label{eq:polarcoordinate}
\begin{split}
&l_{\rm{pixel}} = \sqrt{x^2+y^2}    \\
&\theta = \arcsin(l_{\rm{pixel}}k_{\rm{pixel2mm}}/r)  \\
&\phi = \arctan(x/y)    \\
\end{split}
\end{equation}
However, according to \cite{eyeoptics}, the fovea is not located at the posterior pole of the eyeball but located along the visual axis and the angle between the visual axis and the center axis of the eyeball, which is called the optical axis, is 5 degrees ($\kappa$) as shown in Fig.\ref{fig:2D3Dconvert}(c). Since the visual axis passes through the nodal point, the intersection of the visual axis and the optical axis can be considered to be the nodal point, and the nodal point is located at 16.4 mm ($l_{\rm{nodal}}$) from the posterior pole. Thus, the position of the fovea and the angle ($\kappa_2$) from the center axis is calculated as follows. 
\begin{equation}\label{eq:foveaconpensate1}
\begin{split}
\left\{ \,
    \begin{aligned}
    & y = \tan(\kappa)(x+(l_{\rm{nodal}}-r))  \\
    & x^2 + y^2 = r^2 \\
    \end{aligned}
\right. \\
\end{split}
\end{equation}
\begin{equation}\label{eq:foveaconpensate2}
\begin{split}
&\kappa_2 = \arcsin(y/r)    \\
\end{split}
\end{equation}
$\kappa_2$ is obtained as 6.77 deg.  
The target position ($\vb{p}_{\rm{target3D2}}$) from the posterior pole is obtained by rotating the target position ($\vb{p}_{\rm{target3D}}$) from Eq.\ref{eq:polarcoordinate} around the X axis. The rotation matrix around the X-axis is defined as $rotx$.
\begin{equation}\label{eq:foveaconpensate3}
\begin{split}
&\vb{p}_{\rm{target3D2}} = rotx(\kappa_2)\vb{p}_{\rm{target3D}}   \\
\end{split}
\end{equation}

\subsection{Function2. 6D detection of the eye and trocar }\label{sec: eyeposedetection}
The position and orientation of both the trocar and eyeball must be identified. Recently,  different accurate methods for trocar pose estimation have been developed in our group, achieving an average error of 3.06 $\pm$2.36 degrees\cite{colibri5,colibridoc}. Several approaches for detecting eye gaze have been explored \cite{eyegazetrack1}. Utilizing these methods, the eye posture and trocar 6D pose are estimated. 

\subsection{Function3. Optimization of the eye posture }\label{sec: eyepostureproposal}
\begin{figure}[t]
    \centering
    \vspace{7pt}
    \includegraphics[width=0.7\linewidth]{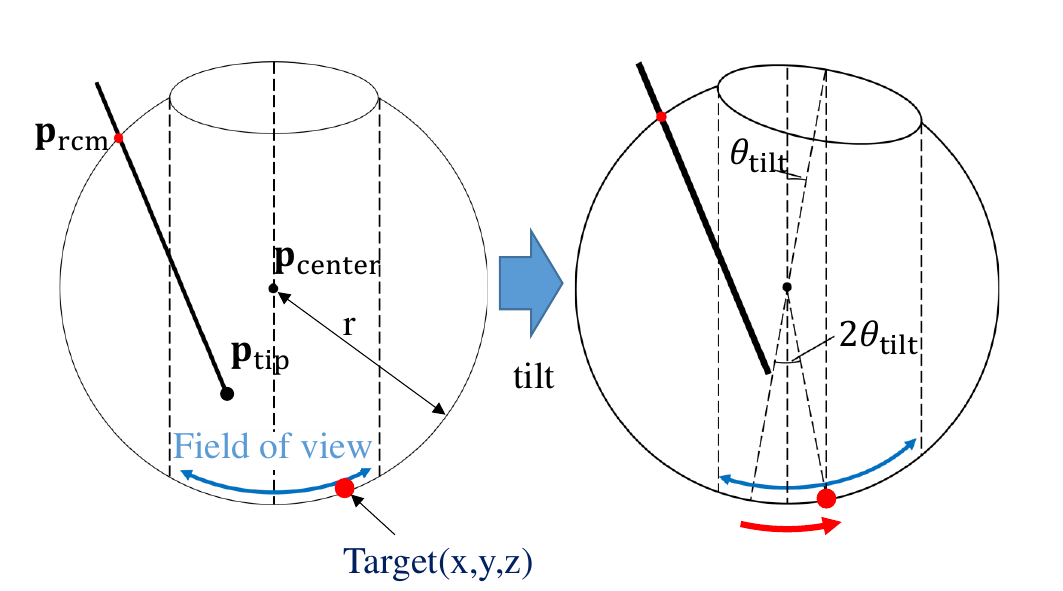}
    \caption{Function3. Optimization of the eye posture. It shows the the shift of the field of view (FoV). When the eyeball tilted $\theta_{\rm{tilt}}$, the FoV shifted $-2\theta_{\rm{tilt}}$ from the posterior pole.} 
    \label{fig:eyeballtilt}
\end{figure}

In this function, the tilting angle of the eyeball $\alpha$ and $\beta$ around the X and Y axis is calculated to move the center of the visible area to the center of the target. Fig.~\ref{fig:eyeballtilt} explains when the eyeball is tilted $\theta_{\rm{tilt}}$, the center of the field view will be rotated the opposite direction and shifted $-2\theta_{\rm{tilt}}$ from the initial center of visible area, which is defined as $\vb{p}_{\rm{fov}}=[0, 0, -r]^T$. This point is shifted to $\vb{p}_{\rm{fov1}}$ and then shifted to $\vb{p}_{\rm{fov2}}$ by the rotation around the X and Y axis.
$\vb{p}_{\rm{fov2}}$ is calculated using the rotation matrix, where $rotx$ and $roty$ are the rotation matrix around the X and Y axis, respectively.
\begin{equation}\label{eq:RCMshift}
\begin{split}
    \vb{p}_{\rm{fov2}} &= roty(-2\beta)rotx(-2\alpha)\vb{p}_{\rm{fov}}\\
    &= \begin{bmatrix}
    r\sin2\beta\cos2\alpha \\
    -r\sin2\alpha \\
    -r\cos2\alpha\cos2\beta \\
    \end{bmatrix}\\
\end{split}
\end{equation}
When the target position is defined as $\vb{p}_{\rm{target}} = [x, y, z]^T$, $\vb{p}_{\rm{fov2}}$ will be equal with the target. $\alpha$ and $\beta$ is obtained. 
\begin{equation}\label{eq:insertionvector}
\begin{split}
    &\alpha = \tfrac{1}{2}\arcsin(-y/r) \\
    &\beta = \tfrac{1}{2}\arcsin(x/(r\cos2\alpha)) \\
\end{split}
\end{equation}
The maximum tilting angle is defined as $\pm$10 deg to minimize the movement of the trocars by tilting the eyeballs. 

\subsection{Function4. Trocar selection and initial configuration}\label{sec: trocarselection}
This function selects the best trocar position to insert the instrument into the eye. Thereafter, the initial tilting angle of the robot and the configuration are also calculated. 

\subsubsection{Trocar selection}
As shown in Fig.\ref{fig:trocarselection}(b), the robot changes the position along the X-axis. Therefore, the best trocar is defined as the closest trocar to the target along the X-axis after tilting the eyeball.
When the initial positions of the trocar and the target are defined as $\vb{p}_{\rm{trocar}}$ and $\vb{p}_{\rm{target}}$, the positions after tilting the eyeball, $\vb{p}_{\rm{trocar2}}$ and $\vb{p}_{\rm{target2}}$ 
 (Fig.\ref{fig:trocarselection}(c)), can be calculated as follows.
\begin{equation}\label{eq:trocarshift}
\begin{split}
    \vb{p}_{\rm{trocar2}} &= roty(\beta)rotx(\alpha)\vb{p}_{\rm{trocar}}\\
    \vb{p}_{\rm{target2}} &= roty(\beta)rotx(\alpha)\vb{p}_{\rm{target}}\\
\end{split}
\end{equation}
There are three trocars, and all the positions after tilting are calculated, and the closest one along the X-axis is selected.
\subsubsection{Initial tilting angle}\label{sec: initialtilt}
The initial tilting angle ($\theta_{\rm{ini}}$) shown in Fig.\ref{fig:strategy}(c) is obtained from $\vb{v}_{\rm{trocar2target}}$. 
\begin{equation}\label{eq:trocar2target}
\begin{split}
    \vb{v}_{\rm{trocar2target}} &= \vb{p}_{\rm{target2}} - \vb{p}_{\rm{trocar2}}\\
\end{split}
\end{equation}
Considering projecting $\vb{v}_{\rm{trocar2target}}$ into 
the YZ plane, $\vb{v}_{\rm{rcm2target}}'$ (Fig.\ref{fig:trocarselection}(d)), $\theta_{\rm{ini}}$ can be calculated as follows. 
\begin{equation}\label{eq:RCMandFOV1}
\begin{split}
    &\theta_{\rm{ini}} = \arctan(\frac{v_{\rm{rcm2target},z}'}{v_{\rm{rcm2target},y}'})\\
\end{split}
\end{equation}

\subsubsection{Initial configuration}\label{sec:initialconfig}
\begin{figure}[t]
    \centering
    \vspace{7pt}
    \includegraphics[width=1.0\linewidth]{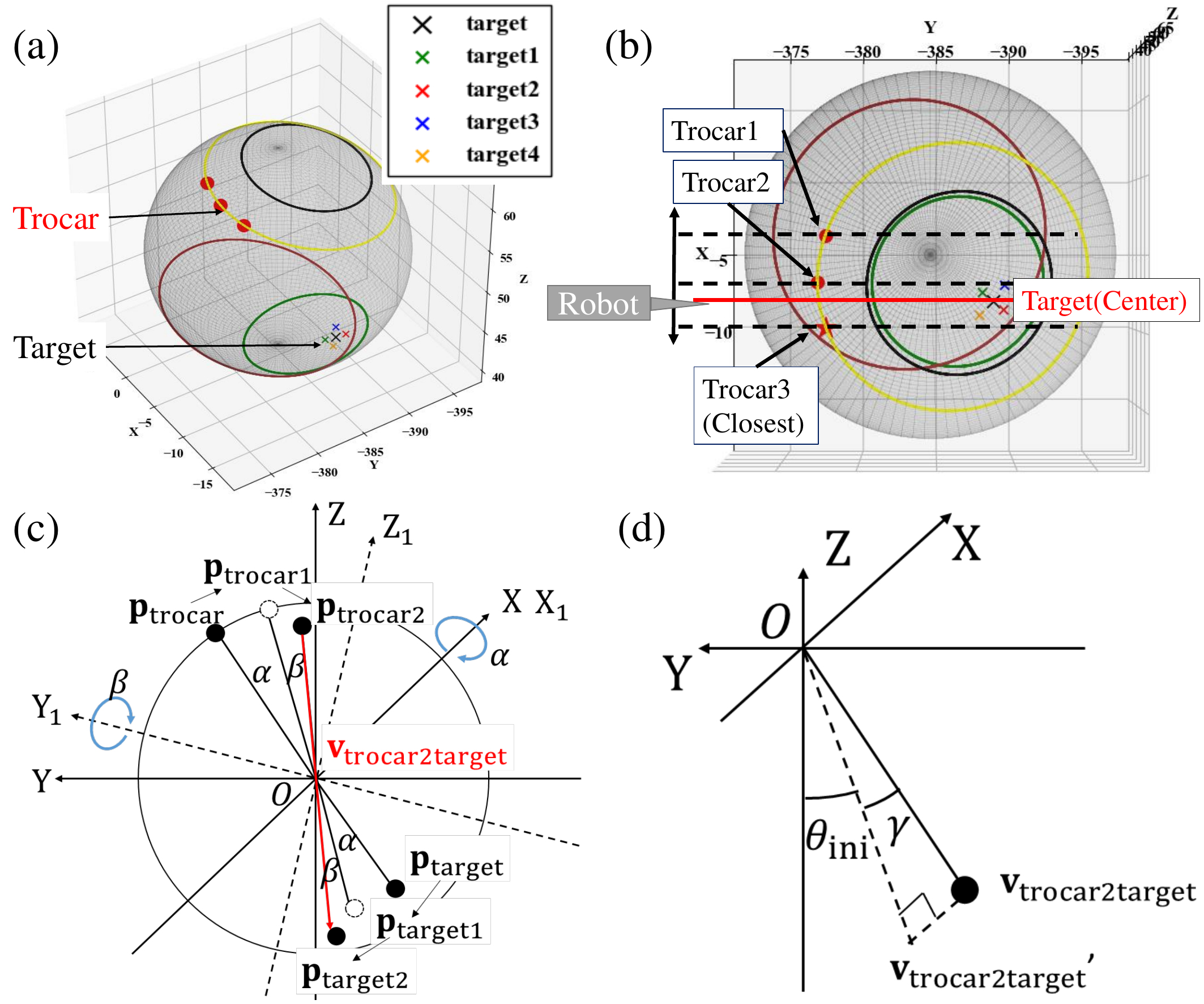}
    \caption{Trocar selection and initial tilting angle. (a) 3D model of the eyeball with trocar and trocar after tilting. (b) Trocar selection. The closest trocar along the X-axis to the center of the target is selected. (c) Movement of the trocar and the target by the eye rotation. (d) Schematic diagram of $\vb{v}_{\rm{trocar2target}}$. $\vb{v}_{\rm{trocar2target}}'$ is the projected vector of $\vb{v}_{\rm{trocar2target}}$ into the YZ plane. } 
    \label{fig:trocarselection}
\end{figure}

In the previous section, the rotation around the X-axis was adjusted, while in this section, the rotation around the Y-axis is refined by optimizing the initial configuration of the surgical robot.
This rotating angle ($\gamma$) is the angle between $\vb{v}_{\rm{trocar2target}}$ and $\vb{v}_{\rm{rcm2target}}'$ (Fig.\ref{fig:trocarselection}(d)) and this is obtained as follows. 
\begin{equation}\label{eq:RCMandFOV2}
\begin{split}
    &\gamma = \arccos(\frac{\vb{v}_{\rm{rcm2target}}\cdot\vb{v}_{\rm{rcm2target}}'}{|\vb{v}_{\rm{rcm2target}}||\vb{v}_{\rm{rcm2target}}'|})\\
\end{split}
\end{equation}
This function is responsible for moving the center of the robot's working angle to $\gamma$. 
The developed method is limited to our robot and is therefore described in Sec.\ref{sec:exfunction3-5}.

\subsection{Function5. Target angle calculation}\label{sec:Target angle}
\begin{figure}[b]
    \centering
    \vspace{-5pt}
    \includegraphics[width=0.8\linewidth]{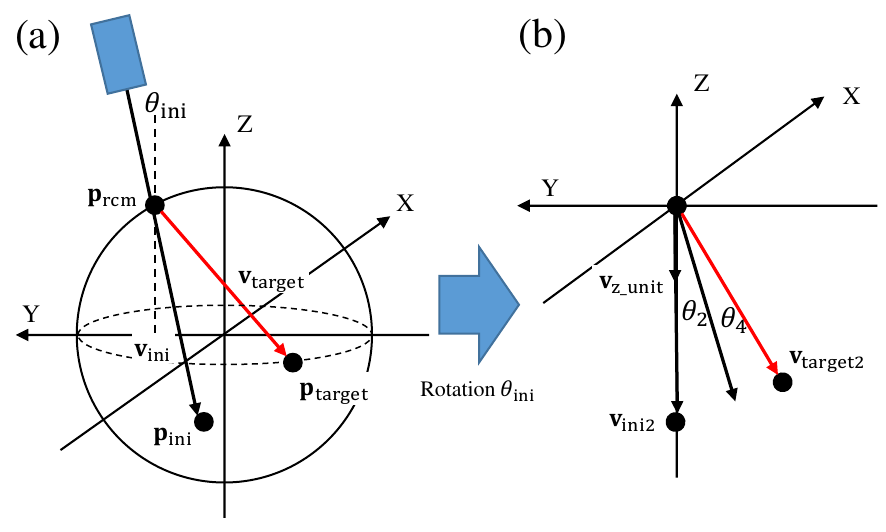}
    \caption{Function5: Target angle calculation. $\vb{v}_{\rm{ini}}$ shows the instrument vector at the initial position. $\vb{v}_{\rm{ini}}$ is the RCM point to target vector. (b) shows the vector after rotating $\vb{v}_{\rm{ini}}$ around the $x$ axis.} 
    \label{fig:targetangle}
    \vspace{12pt}
\end{figure}

In this function, we assume pointing multiple targets (Fig.\ref{fig:trocarselection}(a)). After inserting the instrument into the eyeball, the surgical robot rotates the instrument $\theta_2$ and $\theta_4$ (Fig.\ref{fig:kinematics}) around the X and Y axes and moves it in the depth (Z) direction. These values are obtained in this function. In order to calculate the target angle of the robot, the initial vector ($\vb{v}_{\rm{ini}}$) and the target vector ($\vb{v}_{\rm{target}}$) after tilting the eyeball (Fig.\ref{fig:targetangle}) are obtained. $\vb{v}_{\rm{target}}$ can be obtained in same way as Sec.\ref{sec: trocarselection}.
The initial vector ($\vb{v}_{\rm{ini}}$) represents the vector of the instrument without any movement after the insertion. Therefore, this vector is a rotated vector in the Z-axis direction $\theta_{\rm{ini}}$ around the X-axis. 
In order to simplify, $\vb{v}_{\rm{ini}}$ and  $\vb{v}_{\rm{target}}$ are rotated $\theta_{\rm{ini}}$ in reverse direction as shown in Fig.\ref{fig:targetangle}(b). Then, the rotated vector of $\vb{v}_{\rm{ini}}$,  $\vb{v}_{\rm{ini2}}$, will be in Z axis direction. 
The rotated vector $\vb{v}_{\rm{target2}}$ and $\vb{v}_{\rm{ini2}}$ are obtained as follows. 
\begin{equation}\label{eq:reverserotate}
\begin{split}
    &\vb{v}_{\rm{target2}} = rotx(\theta_{\rm{ini}})\vb{v}_{\rm{target}}\\
    &\vb{v}_{\rm{ini2}} = x\cdot \vb{v}_{\rm{z-unit}}   \\
\end{split}
\end{equation}
where $x$ is the variable and $\vb{v}_{\rm{z-unit}} = [0, 0, -1]^T$. $\vb{v}_{\rm{target2}}$ can also be obtained with $\theta_2$ and $\theta_4$ as follows.
\begin{equation}\label{eq:targetangle}
\begin{split}
    \vb{v}_{\rm{target2}} &= k\cdot rotx(\theta_4)roty(\theta_2)\vb{v}_{\rm{z-unit}}  \\
    &= k\cdot \begin{bmatrix}
    -\sin \theta_2 \\
    -\cos\theta_2\sin\theta_4 \\
    -\cos\theta_2\cos\theta_4 \\
    \end{bmatrix}\\
\end{split}
\end{equation}
$k$ represents the length of $\vb{v}_{\rm{target2}}$. From Eq.\ref{eq:reverserotate} and Eq.\ref{eq:targetangle}, $\theta_2$ and $\theta_4$ are derived.
\begin{equation}\label{eq:finalangle}
\begin{split}
    &k = |\vb{v}_{\rm{target2}}| \\
    &\theta2 = \arctan(v_{\rm{target2},y}/v_{\rm{target2},z})    \\
    &\theta4 = \arcsin(-v_{\rm{target2},x}/k)  \\
\end{split}
\end{equation}
The depth movement is obtained by Eq.\ref{eq:accessibility}, $\theta_2$ and $\theta_4$.

\section{Experimental Evaluation}
\subsection{Experimental setup}
\begin{figure}[b]
    \centering
    \vspace{-10pt}
    \includegraphics[width=0.8\linewidth]{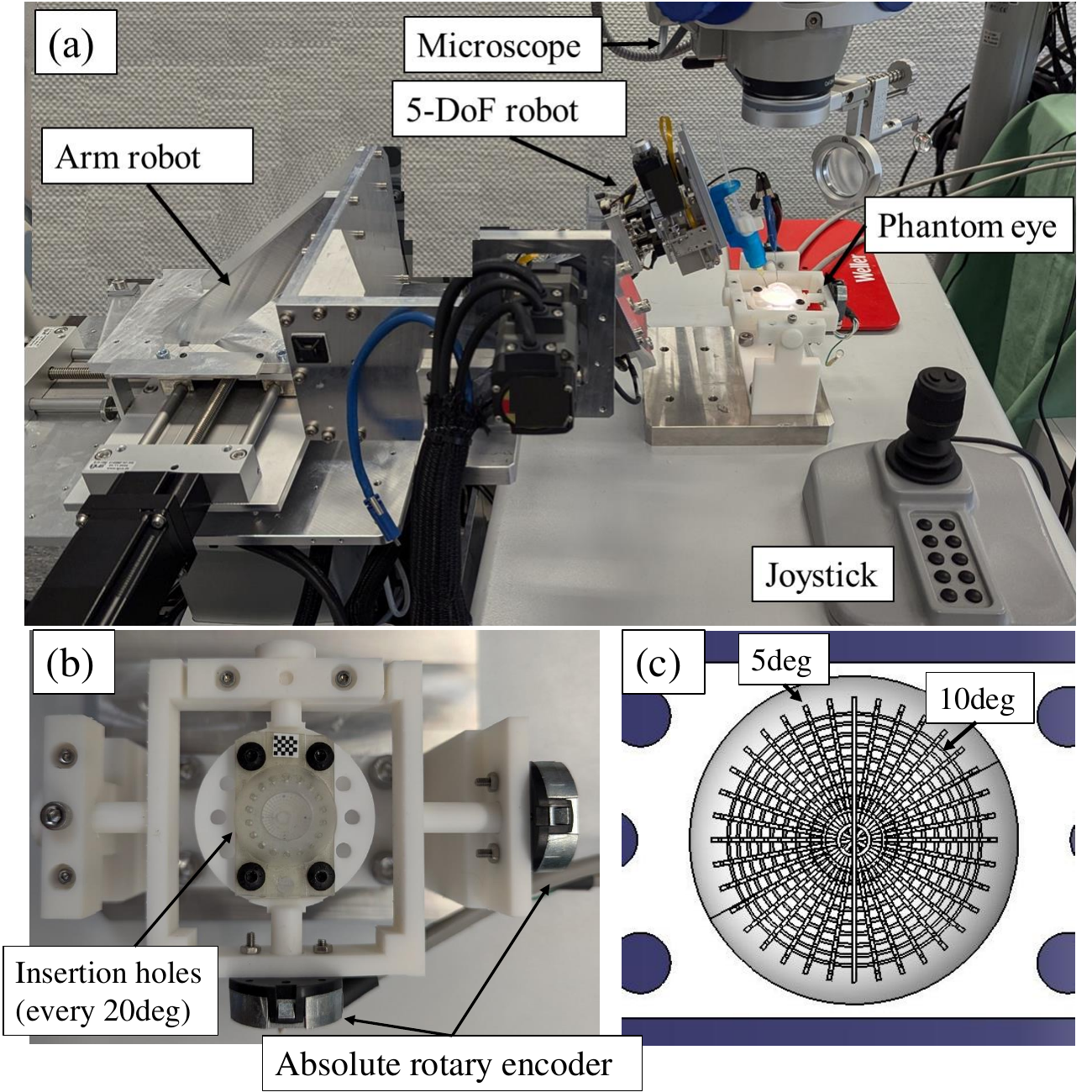}
    \caption{(a) Experimental setup. (b)The eye model can be rotated around the X and Y axis, with absolute encoders mounted at the end of each axis. It has multiple insertion holes in the $20$ degrees step. The eye model can be replaced by other phantom eye. (c)The bottom part of the eye model has grooves on the inner side, which enables the measurement of the position of the fundus. The radial grooves are every $10$deg. Concentric grooves are every $5$deg.} 
    \label{fig:phantomeye}
    \vspace{12pt}
\end{figure}

The phantom eye model is developed with the 3D printer as shown in Fig.~\ref{fig:phantomeye}(b). The bottom part of the eyeball has a grid made of grooves (Fig.~\ref{fig:phantomeye}(c)), and this visualizes the position on the retina. The eye model can rotate around the X and Y axes, and the absolute encoders (AMT222B-V, CUI Devices) are mounted on both axes. The encoder's resolution is 14 bits, and the accuracy is 0.2 degrees. The center of the eye model is designed to keep the position with the different postures. The top part of the eye model contains holes in every $20$ degree to enable varying insertion points. The 3D-printed jig is also designed to keep the same eye model position for all experiments. 

\subsection{Experimental method and Result}
Utilizing this eye phantom, the eye posture and trocar pose (Function 2), as discussed in Sec.\ref{sec: eyeposedetection}, are determined based on encoder data and the known geometries of the phantom.
At first, the functions1 and function3-5 are evaluated separately, and then, the whole workflow is evaluated.
\subsubsection{Function1}\label{sec: ex-function1}
\begin{figure}[t] 
    \centering
    \vspace{5pt}
    \includegraphics[width=0.9\linewidth]{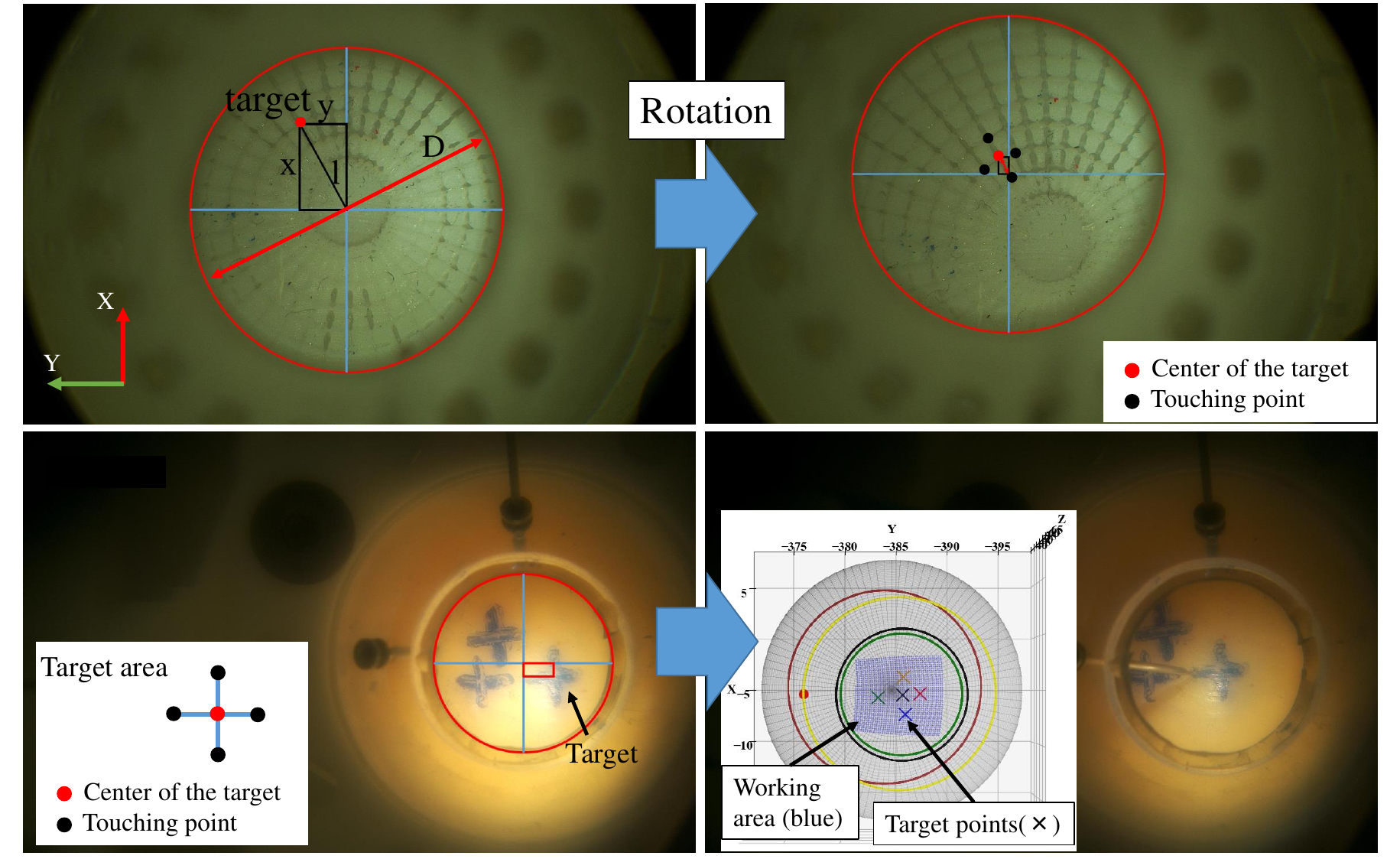}
    \caption{Validation of function1,3. The upper row is the grid phantom eye, and the lower row is the silicon phantom eye. The target and the working area after rotating the eyeball is also simulated. } 
    \label{fig:experiment}
\end{figure}

Since taking the diagnostic image of the phantom eye is hard, the image right in front of the phantom eye is used, as shown in Fig.\ref{fig:experiment}, instead of the diagnostic image. The diagnostic image is a projected image of the fundus. Therefore, we consider that to be a substitute.
In this experiment, the diameter of the cornea is used as a reference to convert pixels to mm. An ellipse is manually fitted, and the center is also detected (Fig.\ref{fig:experiment}). The target position ($x$, $y$) is measured, and the 3D position of the target is obtained by Sec.\ref{sec: targetposcal}. The 3D reconstruction is done without considering the deviation of the visual axis and optical axis. In each experiment, five target points are defined, and all positions are calculated. The center of the target ($\theta_{\rm{center}}$, $\phi_{\rm{center}}$) is defined at the intersection of the grid and the rest four targets are defined at ($\theta_{\rm{center}}\pm$ 5 deg, $\phi_{\rm{center}}\pm$ 10 deg). This experiment is repeated four times.
As a result, the error is -1.65 $\pm$ 0.50 deg ($\theta$) and 0.81 $\pm$ 1.44 deg ($\phi$). 

\subsubsection{Function3-5} \label{sec:exfunction3-5}
Before explaining the experiment, the method of initial configuration in Sec.\ref{sec: trocarselection} is described. In our robot, the rotational joint around the Y axis is $\theta_2$, and this rotational angle is determined by the two linear actuators along the X axis in Fig\ref{fig:kinematics}(b). Our robot can adjust its working angle by changing its initial position. 
The default initial position is the center of the stroke, and the stroke is $\pm$ 13.5 mm. To obtain the best initial position, the initial position is changed from -13.5 to +13.5 in 0.1 mm step, and the working angle and the center of the working angle are calculated each time, and then the position where the center of the working angle is the closest to $\gamma$ is selected.

In this experiment, the target position is directly defined in polar coordinates. First, the tilt angle is calculated by Function 3, and the phantom eye is manually tilted to within $\pm$ 0.1 deg of the obtained angle. The deviation between the target and the center of the ellipse fitted to the cornea is measured and compared with the calculation. 

Next, the initial configuration is done by function4, and the instrument is inserted, and the four target points are manually pointed with the joystick.
When the instrument reaches the target, the joint angle of the robot is collected and compared with the calculated value.
The whole process was repeated four times. As a result, the error of the function3 was 0.45 $\pm$ 0.21 mm, and the error of the function5 was 0.23 $\pm$ 0.99 deg ($\theta_2$), -0.91 $\pm$ 1.07 deg ($\theta_4$), and 0.68 $\pm$ 0.87 mm (Z).

To evaluate the initial configuration in function4, the working angle of $\theta_2$ is also measured. When the center of the target position is defined at (150 deg, 140 deg) on the fundus as an example, $\gamma$ and the initial position is obtained at -4.19 deg and 5.90 mm respectively. The working angle ($\theta_2$) was measured before and after adjusting the initial position. Before the adjustment, the working angle was from -9.86 deg to 9.51 deg and the center of the working angle was -0.18 deg. After the adjustment, the working angle was from -13.88 deg to 5.39 deg and the center of the working angle was -4.25 deg. It is proved that the center of the working angle is shifted to the target position without decreasing the entire working angle by adjusting the initial position. 

\subsubsection{Function1,3-5}\label{sec:exfunction1,3-5}

Four trials are conducted to evaluate the whole proposed workflow, and the error is 0.13 $\pm$ 1.65 deg ($\theta_2$), -1.40 $\pm$ 1.13 deg ($\theta_4$), and 1.80 $\pm$ 1.51 mm (Z).  
\subsubsection{Silicon phantom eye (Function1,3-5)}\label{sec:exphantomeye}
The experiment with a silicon phantom eye is also conducted (Fig.\ref{fig:experiment}). Cross marks are drawn on the fundus as targets. The intersection point of the cross mark is defined as the center of the target, and the edge of the cross mark is defined as a touching point. The whole process was executed as before and repeated four times. The trocars are manually planted each time. The error is 1.07 $\pm$ 2.98 deg ($\theta_2$), -0.90 $\pm$ 1.42 deg ($\theta_4$), and 2.05 $\pm$ 1.67 mm (Z).

\section{Discussion and Future Work}
\begin{figure}[t]
    \centering
    \vspace{5pt}
    \includegraphics[width=1\linewidth]{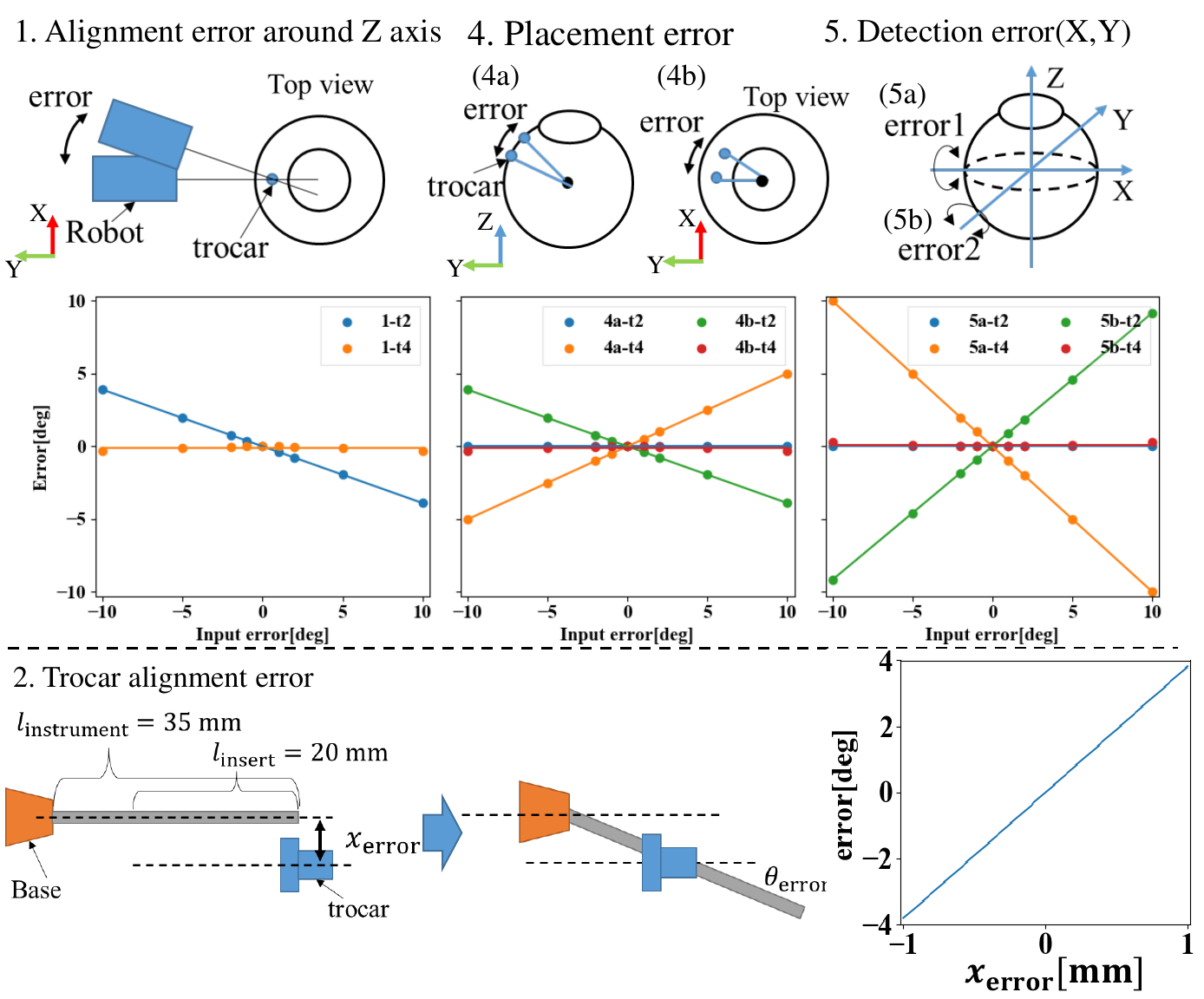}
    \caption{ The analysis of the error sources. In the plots, t2 means $\theta_2$ and t4 means $\theta_4$.} 
    \label{fig:errorsourse}
\end{figure}

We demonstrated the effectiveness of our proposed method through various experiments, despite a certain range of errors. Since the acceptable error depends on factors such as the robot's working angle and the target area's size, establishing a general requirement seems challenging for acceptable error. Hence, discussing the potential error sources is a lead for future research.
Firstly, regarding the error source of the experiment in Sec.\ref{sec:exfunction3-5}, the 3D position of the target is directly defined, and the position of the insertion point and the pose of the eye model is known. Thus, the error sources are limited compared to actual surgery and are enumerated below.
\begin{enumerate}
  \item Alignment error between the robot and the eyeball around the Z axis
  \item Alignment error between the instrument and the trocar
  \item Kinematic error of the robot (attaching the instrument)
\end{enumerate}
In the actual surgery, the error will be added as follows.
\begin{enumerate}[start=4]
  \item Placement and detection error of the trocar
  \item Tilting and detection error of the eye pose
  \item 3D reconstruction error (function1)
  \item Levelness of microscope image plane.
\end{enumerate}
The effect of the error(1), (2), (4), and (5) are analyzed as shown in Fig.\ref{fig:errorsourse}. For simplicity, the target points of the simulation for error(1), (4), and (5) are defined at ($\theta$, $\phi$) = (170 deg, 0 deg), (170 deg, 90 deg), (170 deg, 180 deg), (170 deg, 270 deg) and the center of the target is (180 deg, 0 deg). The input error is changed in [-10, -5, -2, -1, 0, 1, 2, 5, 10](deg), and the approximate straight line is fitted by the least-squares method. Considering the experiment in Sec.\ref{sec:exfunction3-5}, the both joint, $\theta_2$ and $\theta_4$ had error. From the analysis, the error(1) causes the error only in $\theta_2$. Therefore, the error is also caused by the error(2) or (3). Since the alignment error between the instrument and the trocar (2) is hard to simulate, it is simplified to the case that the parallel instrument and trocar are not aligned in a straight line and are separated by an error margin ($x_{\rm{error}}$) as shown in Fig.\ref{fig:errorsourse}-2. In this case, the instrument bends from the base of an instrument. Thus, the error ($\theta_{\rm{error}}$) is obtained as follows.
\begin{equation}\label{eq:error2}
\begin{split}
    &\theta_{\rm{error}} = \arctan(\frac{x_{\rm{error}}}{l_{\rm{instrument}}-l_{\rm{insert}}}) \\
\end{split}
\end{equation}
where $l_{\rm{instrument}}$ and $l_{\rm{insert}}$ are the instrument length and the insertion depth. $l_{\rm{instrument}}$ and $l_{\rm{insert}}$ are fixed at 35 mm and 20 mm and $x_{\rm{error}}$ is changed from -1.0 to 1.0. As shown in Fig.\ref{fig:errorsourse}(2), it is almost linear, and the error will be 1.91 deg when $x_{\rm{error}} = 0.5$ mm. This error can be occurred both in $\theta_2$ and $\theta_4$. Since the operator manually aligns the instrument, it is quite possible that this error will occur.

As for the experiment in Sec.\ref{sec:exfunction1,3-5}, the error was similar to that of the experiment in Sec.\ref{sec:exfunction3-5}. Hence, the error due to function 1 (error 6) is considered insignificant. 
In practice, however, the eyeball is not a perfect sphere, and the angle ($\kappa$) between the visual axis and optical axis varies from person to person. Furthermore, errors occur in the conversion from pixel to actual length. Therefore, a more accurate conversion method is required.

The experiments with silicon phantom eyeballs had larger errors than others (Sec:\ref{sec:exphantomeye}). The difference from the other experiment is the placement of the trocar. This error can be expressed as error in roll (Fig.\ref{fig:errorsourse}(4a)) and yaw (Fig.\ref{fig:errorsourse}(4b)). From the plots, the error in roll (4a) leads to the error in $\theta_4$, and the error in yaw (4b) leads to the error in $\theta_2$. As for the error in yaw (4b), there is usually a thorn at the end of the trocar holder for marking. Therefore, this error is smaller than the error in roll (4a), which is proved in Sec.\ref{sec:exphantomeye}. Furthermore, the instrument and trocar alignment is more difficult than in the previous experiment. This is also the reason why the error is larger than other experiments. In order to reduce the error, accurate eye pose detection and image guidance using AR and jigs for the alignment are required, and these will be the future work. In addition, optimization of trocar placement with AR guidance eliminates the need for three trocars and the initial positional configuration of the robot in Sec.\ref{sec:initialconfig}.

In the actual surgery, the eye posture is detected by image processing. Analysis of this error shows that the detection error of eye posture causes a larger error in the robot joint than other errors (Fig.\ref{fig:errorsourse}(5)). Therefore, a highly accurate eye posture detection method is required.
The microscopic image plane's levelness is also the visible area's error source. Hence, the calibration method to optimize the level of the microscopic image plane will also be required in the future.

\section{Conclusion}
This paper introduces a workflow for optimizing the eye posture, robot tilt, and insertion points depending on the target's position. The entire framework is explained and validated through experiments involving an eye model and a robot. The error of this workflow (Sec.\ref{sec:exfunction1,3-5}) was 0.13 $\pm$ 1.65 deg ($\theta_2$), -1.40 $\pm$ 1.13 deg ($\theta_4$), and 1.80 $\pm$ 1.51 mm (Z). 
The error sources are also discussed. The alignment of the trocar and instrument, as well as the placement of the trocar, etc, are identified as factors influencing the accuracy of the final target angle and the position. Therefore, highly accurate eye and trocar posture detection methods and methods for accurately inserting instruments into trocars are required in the future.

\clearpage

\bibliographystyle{ieeetr}
\bibliography{ref}

\begin{thebibliography}{10}

\bibitem{10611207}
A.~Alikhani, S.~Inagaki, S.~Dehghani, M.~Maier, N.~Navab, and M.~A. Nasseri, ``Envibroscope: Real-time monitoring and prediction of environmental motion for enhancing safety in robot-assisted microsurgery,'' in {\em 2024 IEEE International Conference on Robotics and Automation (ICRA)}, pp.~8202--8208, 2024.

\bibitem{KUleventowardsaclinically}
A.~Gijbels, K.~Willekens, L.~Esteveny, P.~Stalmans, D.~Reynaerts, and E.~Vander~Poorten, ``Towards a clinically applicable robotic assistance system for retinal vein cannulation,'' in {\em 2016 6th IEEE International Conference on Biomedical Robotics and Biomechatronics (BioRob)}, pp.~284--291, 2016.

\bibitem{sixdegree1993}
K.~Grace, J.~Colgate, M.~Glucksberg, and J.~Chun, ``A six degree of freedom micromanipulator for ophthalmic surgery,'' in {\em [1993] Proceedings IEEE International Conference on Robotics and Automation}, pp.~630--635 vol.1, 1993.

\bibitem{gerber2020advanced}
M.~J. Gerber, M.~Pettenkofer, and J.-P. Hubschman, ``Advanced robotic surgical systems in ophthalmology,'' {\em Eye}, vol.~34, no.~9, pp.~1554--1562, 2020.

\bibitem{rahimy2013robot}
E.~Rahimy, J.~Wilson, T.~C. Tsao, S.~Schwartz, and J.~P. Hubschman, ``{Robot-assisted intraocular surgery: development of the IRISS and feasibility studies in an animal model},'' {\em Eye}, vol.~27, no.~8, pp.~972--978, 2013.

\bibitem{nasseri2013introduction}
M.~A. Nasseri, M.~Eder, S.~Nair, E.~C. Dean, M.~Maier, D.~Zapp, C.~P. Lohmann, and A.~Knoll, ``{The introduction of a new robot for assistance in ophthalmic surgery},'' in {\em Eng. Med. Biol. Soc. (EMBC), 2013 35th Annu. Int. Conf. IEEE}, pp.~5682--5685, IEEE, 2013.

\bibitem{10340955}
A.~Alikhani, S.~Oßner, S.~Dehghani, B.~Busam, S.~Inagaki, M.~Maier, N.~Navab, and M.~A. Nasseri, ``Rcit: A robust catadioptric-based instrument 3d tracking method for microsurgical instruments in a single-camera system,'' in {\em 2023 45th Annual International Conference of the IEEE Engineering in Medicine \& Biology Society (EMBC)}, pp.~1--5, 2023.

\bibitem{he2012toward}
X.~He, D.~Roppenecker, D.~Gierlach, M.~Balicki, K.~Olds, P.~Gehlbach, J.~Handa, R.~Taylor, and I.~Iordachita, ``Toward clinically applicable steady-hand eye robot for vitreoretinal surgery,'' in {\em ASME International Mechanical Engineering Congress and Exposition}, vol.~45189, pp.~145--153, American Society of Mechanical Engineers, 2012.

\bibitem{KUlevensetupandmethod}
J.~Smits, D.~Reynaerts, and E.~V. Poorten, ``Setup and method for remote center of motion positioning guidance during robot-assisted surgery,'' in {\em 2019 IEEE/RSJ International Conference on Intelligent Robots and Systems (IROS)}, pp.~1315--1322, 2019.

\bibitem{Koyama}
Y.~Koyama, M.~M. Marinho, and K.~Harada, ``Vitreoretinal surgical robotic system with autonomous orbital manipulation using vector-field inequalities,'' in {\em 2023 IEEE International Conference on Robotics and Automation (ICRA)}, pp.~4654--4660, 2023.

\bibitem{weidesign}
W.~Wei, R.~Goldman, N.~Simaan, H.~Fine, and S.~Chang, ``Design and theoretical evaluation of micro-surgical manipulators for orbital manipulation and intraocular dexterity,'' in {\em Proceedings 2007 IEEE International Conference on Robotics and Automation}, pp.~3389--3395, 2007.

\bibitem{weiperformance}
W.~Wei, R.~E. Goldman, H.~F. Fine, S.~Chang, and N.~Simaan, ``Performance evaluation for multi-arm manipulation of hollow suspended organs,'' {\em IEEE Transactions on Robotics}, vol.~25, no.~1, pp.~147--157, 2009.

\bibitem{haorandesign}
H.~Yu, J.-H. Shen, K.~M. Joos, and N.~Simaan, ``Design, calibration and preliminary testing of a robotic telemanipulator for oct guided retinal surgery,'' in {\em 2013 IEEE International Conference on Robotics and Automation}, pp.~225--231, 2013.

\bibitem{faridpooya2022randomised}
K.~Faridpooya, S.~H. van Romunde, S.~S. Manning, J.~C. van Meurs, G.~J. Naus, M.~J. Beelen, T.~C. Meenink, J.~Smit, and M.~D. de~Smet, ``Randomised controlled trial on robot-assisted versus manual surgery for pucker peeling,'' {\em Clinical \& experimental ophthalmology}, vol.~50, no.~9, pp.~1057--1064, 2022.

\bibitem{accessibilityanalysis}
S.~Inagaki, A.~Alikhani, N.~Navab, M.~Maier, and M.~A. Nasseri, ``Analyzing accessibility in robot-assisted vitreoretinal surgery: Integrating eye posture and robot position,'' in {\em 2024 IEEE International Conference on Robotics and Automation (ICRA)}, pp.~9894--9900, 2024.

\bibitem{davincipreoperative1}
M.~Hayashibe, N.~Suzuki, M.~Hashizume, Y.~Kakeji, K.~Konishi, S.~Suzuki, and A.~Hattori, ``Preoperative planning system for surgical robotics setup with kinematics and haptics,'' {\em The International Journal of Medical Robotics and Computer Assisted Surgery}, vol.~1, no.~2, pp.~76--85, 2005.

\bibitem{davincipreoperative2}
A.~Trejos, R.~Patel, I.~Ross, and B.~Kiaii, ``Optimizing port placement for robot-assisted minimally invasive cardiac surgery,'' {\em The International Journal of Medical Robotics and Computer Assisted Surgery}, vol.~3, no.~4, pp.~355--364, 2007.

\bibitem{davincipreoperative3}
J.~Cannon, J.~Stoll, S.~Selha, P.~Dupont, R.~Howe, and D.~Torchiana, ``Port placement planning in robot-assisted coronary artery bypass,'' {\em IEEE Transactions on Robotics and Automation}, vol.~19, no.~5, pp.~912--917, 2003.

\bibitem{davincipreoperative4}
L.~Sun and C.~K. Yeung, ``Port placement and pose selection of the da vinci surgical system for collision-free intervention based on performance optimization,'' in {\em 2007 IEEE/RSJ International Conference on Intelligent Robots and Systems}, pp.~1951--1956, 2007.

\bibitem{davincipreoperative5}
H.~Azimian, J.~Breetzke, A.~L. Trejos, R.~V. Patel, M.~D. Naish, T.~Peters, J.~Moore, C.~Wedlake, and B.~Kiaii, ``Preoperative planning of robotics-assisted minimally invasive coronary artery bypass grafting,'' in {\em 2010 IEEE International Conference on Robotics and Automation}, pp.~1548--1553, 2010.

\bibitem{davincipreoperative6}
L.~Adhami and E.~Coste-Maniere, ``Optimal planning for minimally invasive surgical robots,'' {\em IEEE Transactions on Robotics and Automation}, vol.~19, no.~5, pp.~854--863, 2003.

\bibitem{davincipreoperative7}
Z.~Du, W.~Wang, W.~Wang, and W.~Dong, ``Preoperative planning for a multi-arm robot-assisted minimally invasive surgery system,'' {\em Simulation}, vol.~93, no.~10, pp.~853--867, 2017.

\bibitem{trocaroptimization1}
F.~Cursi and P.~Kormushev, ``Pre-operative offline optimization of insertion point location for safe and accurate surgical task execution,'' in {\em 2021 IEEE/RSJ International Conference on Intelligent Robots and Systems (IROS)}, pp.~4040--4047, 2021.

\bibitem{trocaroptimization2}
R.~C.~O. Locke and R.~V. Patel, ``Optimal remote center-of-motion location for robotics-assisted minimally-invasive surgery,'' in {\em Proceedings 2007 IEEE International Conference on Robotics and Automation}, pp.~1900--1905, 2007.

\bibitem{trocaroptimization3}
F.~Cursi, W.~Bai, E.~M. Yeatman, and P.~Kormushev, ``Optimization of surgical robotic instrument mounting in a macro–micro manipulator setup for improving task execution,'' {\em IEEE Transactions on Robotics}, vol.~38, no.~5, pp.~2858--2874, 2022.

\bibitem{PKC-RCM}
A.~Alikhani, S.~Inagaki, J.~Yang, S.~Dehghani, M.~Sommersperger, K.~Huang, M.~Maier, N.~Navab, and M.~A. Nasseri, ``Pkc-rcm: Preoperative kinematic calibration for enhancing rcm accuracy in automatic vitreoretinal robotic surgery,'' {\em IEEE Access}, vol.~11, pp.~103616--103627, 2023.

\bibitem{nasseri2013kinematics}
M.~A. Nasseri, M.~Eder, D.~Eberts, S.~Nair, M.~Maier, D.~Zapp, C.~P. Lohmann, and A.~Knoll, ``Kinematics and dynamics analysis of a hybrid parallel-serial micromanipulator designed for biomedical applications,'' in {\em 2013 IEEE/ASME International Conference on Advanced Intelligent Mechatronics}, pp.~293--299, IEEE, 2013.

\bibitem{virtualfixture}
M.~A. Nasseri, P.~Gschirr, M.~Eder, S.~Nair, K.~Kobuch, M.~Maier, D.~Zapp, C.~Lohmann, and A.~Knoll, ``Virtual fixture control of a hybrid parallel-serial robot for assisting ophthalmic surgery: An experimental study,'' in {\em 5th IEEE RAS/EMBS International Conference on Biomedical Robotics and Biomechatronics}, pp.~732--738, 2014.

\bibitem{houghcircle}
N.~Guil and E.~L. Zapata, ``Lower order circle and ellipse hough transform,'' {\em Pattern Recognition}, vol.~30, no.~10, pp.~1729--1744, 1997.

\bibitem{eyeoptics}
M.~Kaschke, K.-H. Donnerhacke, and M.~S. Rill, ``Optical devices in ophthalmology and optometry: technology, design principles and clinical applications,'' 2013.

\bibitem{fundusimage}
J.~Staal, M.~D. Abr{\`a}moff, M.~Niemeijer, M.~A. Viergever, and B.~Van~Ginneken, ``Ridge-based vessel segmentation in color images of the retina,'' {\em IEEE transactions on medical imaging}, vol.~23, no.~4, pp.~501--509, 2004.

\bibitem{colibri5}
S.~Dehghani, M.~Sommersperger, M.~Saleh, A.~Alikhani, B.~Busam, P.~Gehlbach, I.~Iordachita, N.~Navab, and M.~Ali~Nasseri, ``Colibri5: Real-time monocular 5-dof trocar pose tracking for robot-assisted vitreoretinal surgery,'' in {\em 2024 IEEE International Conference on Robotics and Automation (ICRA)}, pp.~4547--4554, 2024.

\bibitem{colibridoc}
S.~Dehghani, M.~Sommersperger, J.~Yang, M.~Salehi, B.~Busam, K.~Huang, P.~Gehlbach, I.~Iordachita, N.~Navab, and M.~A. Nasseri, ``Colibridoc: an eye-in-hand autonomous trocar docking system,'' in {\em 2022 International Conference on Robotics and Automation (ICRA)}, pp.~7717--7723, 2022.

\bibitem{eyegazetrack1}
Z.~Zhu and Q.~Ji, ``Novel eye gaze tracking techniques under natural head movement,'' {\em IEEE Transactions on Biomedical Engineering}, vol.~54, no.~12, pp.~2246--2260, 2007.

\end{thebibliography}

\end{document}